\documentclass[10pt]{article}

\usepackage{libertine}
\usepackage{amsmath, amssymb, graphicx, url}

\usepackage{tikz}
\usetikzlibrary{shapes.geometric, backgrounds, calc, 3d, mindmap, trees, arrows.meta}

\newcommand{\ttc}{Text2Node}
\newcommand{\sno}{SNOMED CT}

\usepackage{xcolor, colortbl}

\usepackage[british]{babel}
\usepackage{authblk}

\title{\ttc: a Cross-Domain System for Mapping Arbitrary Phrases to a Taxonomy}
\author{Rohollah Soltani}
\author{Alexandre Tomberg}
\affil{Knowtions Research Inc. \\Toronto, Ontario, Canada}
\date{}

\begin{document}

\maketitle

\begin{abstract}
Electronic health record (EHR) systems are used extensively throughout the healthcare domain.
However, data interchangeability between EHR systems is limited due to the use of different coding standards
across systems.
Existing methods of mapping coding standards based on manual human experts mapping,
dictionary mapping, symbolic NLP and classification are unscalable and cannot accommodate large scale EHR datasets.

In this work, we present \ttc, a cross-domain mapping system capable of mapping medical phrases
to concepts in a large taxonomy (such as \sno).
The system is designed to generalize from a limited set of training samples and map phrases
to elements of the taxonomy that are not covered by training data.
As a result, our system is scalable, robust to wording variants between coding systems and
can output highly relevant concepts when no exact concept exists in the target taxonomy.
Text2Node operates in three main stages:
first, the lexicon is mapped to word embeddings;
second, the taxonomy is vectorized using node embeddings;
and finally, the mapping function is trained to connect the two embedding spaces.
We compared multiple algorithms and architectures for each stage of the training,
including GloVe and FastText word embeddings, CNN and Bi-LSTM mapping functions, and node2vec for node embeddings.
We confirmed the robustness and generalisation properties of Text2Node by
mapping ICD-9-CM Diagnosis phrases to \sno\ and by zero-shot training at comparable accuracy.

This system is a novel methodological contribution to the task of normalizing and linking phrases to a taxonomy,
advancing data interchangeability in healthcare.
When applied, the system can use electronic health records to generate an embedding that
incorporates taxonomical medical knowledge to improve clinical predictive models.
\end{abstract}

\section{Introduction}

Electronic heath and medical record (EHR/EMR) systems are steadily gaining in popularity.
Ever more facets of healthcare are recorded and coded in such systems, including
patient demographics, disease history and progression, laboratory test results,
clinical procedures and medications, and even genetics.
This trove of information is a unique opportunity to learn patterns
that can help improve various aspects of healthcare.
However, the sheer number of various coding systems used to encode this clinical information
is a major challenge for anyone trying to analyse structured EHR data.
Even the most widely used coding systems have multiple versions to cater to different regions of the world.
Software developed to analyze one version of the coding system may not be used for another version,
let alone a different coding system.
In addition to the public and well documented coding systems,
a multitude of private coding mechanisms that have no mappings to any public coding systems
are used by insurance companies and certain hospitals.

The efforts to solve this problem range from
the development of mapping dictionaries between coding systems to machine learning driven approaches.
One example of the former is cTAKES \cite{savova2010mayo},
a project that uses annotated lookup tables to map clinical entities to concepts in a controlled vocabulary
such as \sno\ (Systematised Nomenclature of Medicine -- Clinical Terms, \cite{lozano2010experiences}).
The reliance of cTAKES and similar systems on symbolic natural language processing techniques makes them
hard to generalise and scale, especially in view of regular updates and changes to the target vocabulary.
An example of the latter approach is the work of \cite{zhao2018neural},
where word embeddings have been used in a hierarchical structure to annotate and map medical concepts to
a reference taxonomy.
Their method is based on classification learning and limited in its ability to be applied on
controlled vocabularies such as \sno\ due to the large size of these vocabularies and
small number of terms or phrases associated with each concepts.

In this work, we present a system that we call \ttc\ that maps phrases from a medical lexicon
to a concept in a large target taxonomy (we use \sno\ in this paper).
Our system is scalable since it relies on lexicon and taxonomy embeddings that are generated in an unsupervised fashion.
It is robust to changes in the phrase vocabulary,
can output similar concepts in cases when the exact match is not found,
and can map to parts of the taxonomy that were not covered by the training data.
We achieved this result by relying on node embeddings built from the relationships between the concepts in \sno,
and by replacing the classification problem of finding the best concept
with a regression to the nearest node embedding.

Applications of the \ttc\ system include
scalable generation of mapping dictionaries for non-public medical coding systems,
identification and recognition of clinical concepts or named entities in EHRs and EMRs,
and generation of phrases representations that are medical concept aware.

\paragraph{Technical Significance}
We developed \ttc\, as a system that maps phrases from a medical lexicon to concepts in a large target taxonomy.
The system relies on word and node embeddings that are generated in a scalable unsupervised fashion;
it is robust to changes in the vocabulary and presence of out-of-vocabulary words;
it returns highly relevant concepts even in cases where the exact match is not found or does not exist;
and it generalises well as demonstrated by our zero-shot learning experiments.
A key ingredient in achieving these results was the transformation of
the classification task of finding the best concept into a regression to the nearest node embedding problem.

\paragraph{Clinical Relevance}
Our \ttc\ system can be used for generating mapping dictionaries for non-public medical coding systems
to large public taxonomies in a scalable way.
It is robust to changes in wording between coding systems,
it can generalise well from a limited set of training samples,
and it can map to elements of taxonomies that are not covered by training data.
The \ttc\ system can also be used for identification and recognition of clinical concepts and clinical named entities
in electronic medical records.

\section{Motivation}\label{sec:motivation}

A natural approach to mapping an arbitrary coding system to a well documented public taxonomy is to start with
the phrases that describe each code.
Since the coding system must be understandable to its human users, the phrases describing each code must provide
enough information for them to be able to use it.

\paragraph{Lexical representation}
Since the seminal work of \cite{mikolov2013efficient} on Word2Vec, word embeddings have been widely used
to capture the \emph{semantic meaning} of words, phrases and even sentences.
For example, word embeddings have been successfully applied in clinical settings to
information retrieval, named entity recognition and patient outcome prediction tasks on
unstructured text in EMR narratives.
Specifically, this technique assigns real-valued vectors of a fixed length to individual words
from a collection of documents, called a \emph{corpus}.
This vector representation is such that it captures the semantic relationships between words,
so that synonyms lie close to each other, while unrelated words are far away.
Following Word2Vec,
many algorithms have been developed and used successfully to generate word embeddings,
including GloVe \cite{pennington2014glove} and FastText \cite{bojanowski2017enriching}.

The defining feature of all of word embedding algorithms is the use of contextual interchangeability
as a proxy for relatedness in meaning.
However, this feature can be a problem for the task at hand, especially in the medical setting.
For example, the terms \emph{cold} and \emph{viral respiratory infection} are highly related,
but are not often used interchangeably in the medical context.
The use of contextual interchangeability as a proxy will lead the word algorithms to incorrectly position
the vectors corresponding to these two terms very far from each other in the embedding space.

\paragraph{Taxonomical representation}
In healthcare and biomedical research, the relationships between entities contain valuable information,
because they describe the interactions and causal relationships between diagnosis, medications and procedures,
as well as genetic components. To document the complex relationships,
large databases have been built, including
biomedical knowledge graphs (e.g.
PharmGKB \cite{klein2001integrating},
DrugBank \cite{wishart2006drugbank}
),
ontologies (e.g. Gene Ontology \cite{geneontology2014})
and taxonomies such as
International Statistical Classification of Diseases (ICD), and \sno.

Network topology is used to analyse and represent the network structure of these biomedical databases.
Such analysis requires high computational costs due to
the high dimensionality and sparsity of these databases.
Network embedding technologies provide new effective paradigms to solve the network analysis problem. It converts
network into a low-dimensional space while maximally preserving its structural properties.
Many network embedding algorithms have been developed to embed these graphs into vector spaces
(see for example \cite{su2018network})
and used successfully to predict drug-drug interactions \cite{wang2017predicting} and another paper.

\paragraph{Mapping between representations}
Since contextual interchangeability is not a good metric of medical relatedness,
word embeddings cannot be directly used to map between coding systems.
However, they can be used to capture semantic information from phrases that are used to describe such systems.
In contrast, node embeddings generated from concepts in a medical taxonomy are a better representation of
medical relatedness, because they are built from relationships between medical concepts.
In order to bridge the gap between these two embedding spaces, a mapping function is needed.
This mapping function is to operate on the level of vector representations rather than
original phrases and concepts.
This has two important benefits:
these vector spaces are low-dimensional compared to hundreds of thousands of original concepts;
the function learned from embeddings should be more generalisable and is in general easier to train.

\paragraph{Zero-shot learning}
Whenever there is scarcity of supervised data, machine learning models often fail to carry out reliable generalisations.
Obtaining correctly labelled data is often costly and impractical for large datasets.
A practical application of concept embedding is the zero-shot transformation of words and concepts
that were missing in the training data.
It is possible to generalize the mapping function and accurately map unseen concepts,
despite having only a few training examples per concept,
because embedding training in both domains is an unsupervised task.
This is done through nearest neighbour retrieval,
where the closest embedding in the target space is selected according to some similarity metric.

\section{Methods}

A visual representation of the \ttc\ system is provided by Figure~\ref{fig:overview}.
There are three main stages:
first, the lexicon is mapped to word embeddings;
second, the taxonomy is vectorized using node embeddings;
and finally, the mapping function is trained to connect the two embedding spaces.

% System overview figure
\begin{figure}[htbp]
  \centering
  \begin{tikzpicture}[shorten >=3pt, scale=.55]
    % \draw[step=1cm, gray, very thin] (-5.5,-0.5) grid (9.9, 9.9);

    % Word embeddings
    \begin{scope}[shift={(-6,0)}, local bounding box=scope_left]
      \node[align=center] at (1.5, 3.5) {Word embeddings};
      \draw[black, ->] (0, -0.3) -- (0, 3.3);
      \draw[black, ->] (-0.3, 0) -- (3.3, 0);
      \foreach \Point in {(.5, 1), (1.5, .5), (.5, 2.5), (1.5, 2), (2.7, 1.3)}{
          \node[blue] at \Point {\textbullet};
        }
      \begin{pgfonlayer}{background}
        \draw[black!15,fill=black!5,rounded corners=1ex]
        ($(scope_left.south west) + (-.2, -.2)$) rectangle ($(scope_left.north east) + (.2, .2)$);
      \end{pgfonlayer}
    \end{scope}

    % Node embeddings
    \begin{scope}[shift={(6,0)}, local bounding box=scope_right]
      \node[align=center] at (1.5, 3.5) {Node embeddings};
      \draw[black, ->] (0, -0.3) -- (0, 3.3);
      \draw[black, ->] (-0.3, 0) -- (3.3, 0);
      \foreach \Point in {(.5, 1), (1.5, .5), (.5, 2.5), (1.5, 2), (2.7, 1.3)}{
          \node[red] at \Point {\textbullet};
        }
      \begin{pgfonlayer}{background}
        \draw[black!15,fill=black!5,rounded corners=1ex]
        ($(scope_right.south west) + (-.2, -.2)$) rectangle ($(scope_right.north east) + (.2, .2)$);
      \end{pgfonlayer}
    \end{scope}

    % Corpus
    \begin{scope}[shift={(-5, 8.7)}, scale=0.6, every node/.append style={transform shape}, local bounding box=scope_docs]
      \def\corner{0.15in};
      \def\cornerradius{0.02in};
      \def\lwidth{0.02in};
      \def\h{1.1in};
      \def\w{0.85in};
      \def\nline{10};
      \def\iconmargin{0.1in};
      \def\topmargin{0.3in};
      \foreach[count=\i] \filename in {doc4,doc3,doc2,doc1}{
          \coordinate (nw) at ($(-0.20in*\i,-0.15in*\i)$);
          \coordinate (ne0) at ($(nw) + (\w, 0)$);
          \coordinate (ne1) at ($(ne0) - (\corner, 0)$);
          \coordinate (ne2) at ($(ne0) - (0, \corner)$);
          \coordinate (se) at ($(ne0) + (0, -\h)$);
          \filldraw [-, line width = \lwidth, fill=white] (nw) -- (ne1) -- (ne2)
          [rounded corners=\cornerradius]--(se) -- (nw|-se) -- cycle;
          \draw [-, line width = \lwidth] (ne1) [rounded corners=\cornerradius]-- (ne1|-ne2) -- (ne2);
          \node [anchor=north west] at (nw) {\scriptsize \tt \filename};
          \foreach \k in {1,...,\nline}{
              \draw [-, line width = \lwidth, line cap=round]
              ($(nw|-se) + (\iconmargin,\iconmargin) + (0,{(\k-1)/(\nline-1)*(\h - \iconmargin - \topmargin)})$)
              -- ++ ($(\w,0) - 2*(\iconmargin,0)$);
            }
        }
      \begin{pgfonlayer}{background}
        \draw[black!15,fill=black!5,rounded corners=1ex]
        ($(scope_docs.south west) + (-1, -2)$) rectangle ($(scope_docs.north east) + (1, .4)$);
      \end{pgfonlayer}
    \end{scope}

    % Taxonomy
    \begin{scope}[shift={(7, 8)}, scale=0.33, every node/.append style={transform shape}, local bounding box=scope_net]
      \tikzset{every concept/.append style={minimum size=1.5cm, text width=1.5cm}}
      \path[mindmap, concept color=black, text=white]
      node[concept] {}
      [clockwise from=0]
      child[concept color=green!50!black] {
          node[concept] {}
            [clockwise from=-30]
          child { node[concept] {} }
          child { node[concept] {} }
        }
      child[concept color=blue] {
          node[concept] {}
            [clockwise from=-30]
          child { node[concept] {} }
          child { node[concept] {} }
        }
      child[concept color=red] { node[concept] {} }
      child[concept color=orange] { node[concept] {} };
      \begin{pgfonlayer}{background}
        \draw[black!15,fill=black!5,rounded corners=1ex]
        ($(scope_net.south west) + (-.5, -2.5)$) rectangle ($(scope_net.north east) + (.5, 1)$);
      \end{pgfonlayer}
    \end{scope}

    \node[align=left, above] at (scope_docs.south) {Corpus};
    \node[align=left, above] at (scope_net.south) {Taxonomy};
    \draw[very thick, ->] (scope_docs.west) to[out=180, in=180] (scope_left.west);
    \draw[very thick, ->] (scope_net.east) to[out=0, in=0] (scope_right.east);

    \path (scope_left.east) -- coordinate[midway](m) (scope_right.west);
    \node[draw, align=center] (cnn) at (m) {Mapping\\function};
    % \draw[very thick, ->, double] (scope_left.east) -- (cnn);
    \draw[double distance=1, -{Latex[length=10, open]}, line width=1] (scope_left.east) -- (cnn);
    \draw[double distance=1, -{Latex[length=10, open]}, line width=1] (cnn) -- (scope_right.west);

  \end{tikzpicture}
  \caption{\ttc\ System overview}
  \label{fig:overview}
\end{figure}

\subsection{Word Embeddings}

\paragraph{Corpus}
According to \cite{wang2018comparison}, word embeddings trained on a biomedical corpus can capture
the semantic meaning of medical concepts better than embeddings trained on an unspecialised set of documents.
We have thus constructed a large corpus consisting of open access papers from PubMed,
free text admission and discharge notes from the MIMIC-III Clinical Database \cite{johnson2016mimic, goldberger2000components},
narratives from the US Food and Drug Administration (FDA) Adverse Event Reporting System (FAERS),
and a part of the 2010 Relations Challenge from i2b2 \cite{uzuner20112010}.

The documents from those sources were pre-processed using the Stanford CoreNLP pipeline \cite{manning2014stanford}
to split sentences, add spaces around punctuation marks, change all characters to lowercase,
and reformat to one sentence per line.
Finally, all files were concatenated into a single document with 235M sentences and 6.25B words.
This file was used to train both algorithms described below.

\paragraph{Algorithms}
We focused on two algorithms for learning word embeddings: GloVe and FastText.
An important distinction between them is the treatment of words that are not part of the training vocabulary:
GloVe creates a special out-of-vocabulary token and maps all of these words to this token's vector,
while FastText uses subword information to generate an appropriate embedding.

We set the vector space dimensionality to 200 and the minimal number of word occurrences to 10 for both algorithms:
this produced a vocabulary of 3.6M tokens.
All other algorithm-specific settings were left at their default values, since we expected that \ttc\ would
perform just as well with different versions of embeddings.

\subsection{Node Embeddings}

\paragraph{Taxonomy} The taxonomy to which \ttc\ maps phrases can be arbitrary.
In this paper, we focused on the 2018 international version of \sno\
as our target graph $\mathcal{G} = (\mathcal{V}, \mathcal{E})$.
The vertex set $\mathcal{V}$ consists of 392K medical concepts
and the edge set $\mathcal{E}$ is composed of 1.9M relations between the vertices
including \texttt{is\_a} relationships and attributes such as \texttt{finding\_site} and \texttt{due\_to}.

\paragraph{Algorithm}
To construct taxonomy embeddings, we used node2vec method from \cite{grover2016node2vec}.
We started a random walk on the edges from each vertex $v \in \mathcal{V}$ and stopped it after a fixed number of steps
(20 in our application).
All the vertices visited by the walk were considered to be part of the graph neighbourhood $N(v)$ of $v$.
Following the skip--gram architecture from \cite{mikolov2013efficient},
we selected the feature vector assignment function $v \mapsto f_\text{n2v}(v) \in \mathbb{R}^{128}$
by solving the optimisation problem
\[
  f_\text{n2v} = \operatorname*{argmax}_f \sum_{u \in \mathcal{V}} \log \mathbb{P} \big[ N(u) | f(u) \big],
\]
using stochastic gradient descent and negative sampling.

% \paragraph{DNGR}
% Improving on the skip-gram on random walks method from node2vec,
% DNGR (\underline{d}eep \underline{n}eural networks for \underline{g}raph \underline{r}epresentations)
% model constructs a positive pointwise mutual information (PPMI) matrix by using random surfing,
% which can capture more global information than a random walk.
% This probabilistic representation $x_v$ of the neighbourhood $N(v)$ of a vertex $v$ is fed into
% a stacked denoising autoencoder to produce an embedding for the node $v$.
% The encoder $f$ and decoder $g$ parts of the network are trained simultaneously to minimise
% \[
%   f, g = \operatorname*{argmin}_{f, g} \sum_{x \in \text{rows of PPMI}} \big\| x -  g(f(\tilde{x})) \big\|^2,
% \]
% where $\tilde{x}$ is the noisy version of $x$ obtained by randomly setting some (how many is some?) entries of $x$ to $0$.
% The embedding of the node $v$ is then given by $f(x_v)$.

\subsection{Mapping}

\begin{figure}[ht]
  \centering
  \tikzstyle{element}=[rectangle, rounded corners, text centered, text=black, text width=2cm]
  \begin{tikzpicture}
    % \draw[step=1cm, gray, very thin] (-5.5,-.5) grid (5.5, 6.5);

    \node[element, draw=green!70!black, fill=green!30!white, text width=4cm] (phrase) at (0,0) {\footnotesize Input Phrase};

    \foreach \i/\x in {1/-3.75, 2/-1.25, 3/1.25, 4/3.75}{
        \node[element, draw=yellow!70!black, fill=yellow!30!white] (word_\i) at (\x, 2) {\scriptsize Word \i};
        \node[element, draw=blue!70!black, fill=blue!30!white] (emb_\i) at (\x, 3) {\scriptsize Embedding \i};
        \draw[thick, ->] (phrase.north) to[out=90, in=-90] (word_\i.south);
        \draw[thick, ->] (word_\i.north) -- (emb_\i.south);
        \draw[thick, ->] (emb_\i.north) -- ++(0, .75);
      }

    \path[draw] (-5, 4) -- (5, 4) -- (0, 6) -- cycle;
    \node at (0, 5) {\footnotesize Mapping Function};
    \node[element, draw=orange!70!black, fill=orange!30!white, text width=4cm] (node) at (0, 7) {\footnotesize Node Embedding};
    \draw[thick, ->] (0, 6) -- (node.south);

  \end{tikzpicture}
  \caption{
    Mapping procedure:
    the input phrase is split into words that are converted to word embeddings and fed into the mapping function;
    the output of the function is a point in the node embedding space $\mathbb{R}^{128}$.
    }
  \label{fig:general_text2code}
\end{figure}
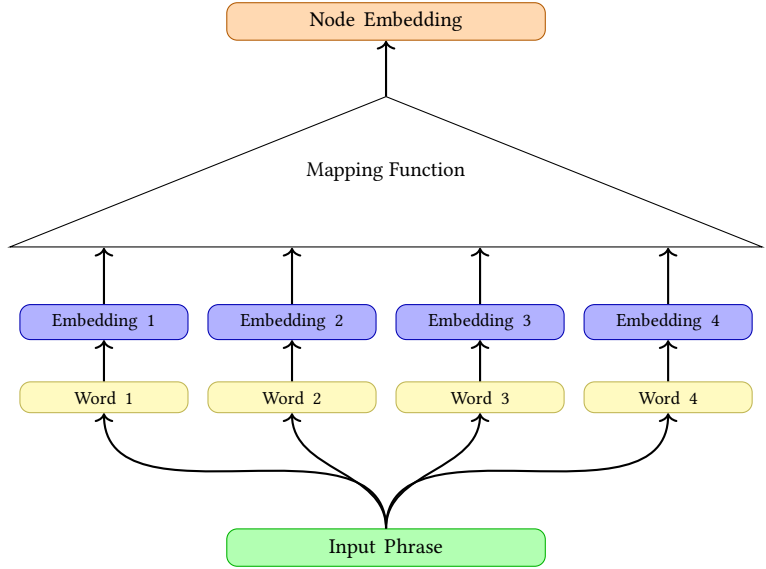

The mapping between phrases and concepts in the target taxonomy was made by associating points in
the node embedding vector space to sequences of word embeddings corresponding to individual words in a phrase.
As illustrated in Figure~\ref{fig:general_text2code},
given a phrase consisting of $n$ words with the associated word embeddings $w_1, \ldots, w_n$,
the mapping function was $m: (w_1, \ldots, w_n) \mapsto p$, where $p \in \mathbb{R}^{128}$ is a point in
the node embedding vector space.
%
% Mapping diagram
To complete the mapping, we found concepts in the taxonomy whose node embeddings were the closest to the point $p$.
We experimented with two measures of closeness in the node embedding vector space $\mathbb{R}^{128}$:
Euclidean $\ell_2$ distance and cosine similarity, that is
\begin{align*}
  \ell_2 \text{ distance }(p,q) &= \|p - q\| = \sqrt{(p - q) \cdot (p - q)}, \\
  \cos \text{ similarity }(p,q) &= \frac{p \cdot q}{\|p\| \|q\|}.
\end{align*}
In some cases, for example to compute the top-$k$ accuracy of the mapping (see Section~\ref{sec:results}),
a list of $k$ closest concepts was used.

\paragraph{Function architectures}
The exact form of the mapping function $m$ can vary.
In this paper, we tested three architectures:
a linear mapping,
a convolutional neural network (CNN), and
a bidirectional long short term memory network (Bi-LSTM).
We padded or truncated all phrases to be exactly 20 words long to represent each phrase by 20 word embeddings
$w_1,\ldots,w_{20} \in \mathbb{R}^{200}$ in order to accommodate all three architectures.

\begin{description}
  \item[Linear]
  We concatenated 20 word embeddings into a single 4000 dimensional vector $w$, and
  the linear mapping was given by $p = m(w) = Mw$ for a $4000 \times 128$ matrix $M$.

  \item[CNN]
  A convolutional neural network \cite{lecun1998gradient} applies convolutional filters of
  different sizes to the input vectors.
  The feature maps produced by the filters are then fed into a pooling layer followed by a projection layer
  to obtain an output of desired dimension.
  Here, we used filters representing word windows of sizes 1, 2, 3, and 5, followed by a maximum pooling layer
  and a projection layer to 128 output dimensions.
  CNN is a non-linear transformation that is more flexible and may be better suited
  to capture complex patterns in the input than a linear mapping method.
  Another useful property of the CNN is its ability
  to learn invariant features regardless of their position in the phrase.

  \item[Bi-LSTM]
  A bidirectional long short term memory network (Bi-LSTM) \cite{schuster1997bidirectional}
  is also a non-linear transformation.
  This type of neural network operates by recursively applying a computation to every element of the input sequence
  conditioned on the previous computed results in both forward and backward directions.
  Bi-LSTM excels at learning long distance dependencies in its input.
  We built a Bi-LSTM to approximate the mapping function $m$ using a single Bi-LSTM cell with 200 hidden units
  followed by a projection layer to 128 output dimensions.

\end{description}

\paragraph{Training}
We gathered training data consisting of phrase--concept pairs from the taxonomy itself.
Because most nodes in \sno\ have multiple phrases describing them (synonyms),
we considered each synonym--concept pair separately for a total of 269K training examples.
To find the best mapping function $m_*$ in each of the three architectures described above,
we solved the supervised regression problem
\[
  m_* = \operatorname*{argmin}_m \sum_{(\text{phrase}, \text{node})}
  \big\|m(\text{phrase}) - f_\text{n2v}(\text{node})\big\|^2_{\ell_2},
\]
using the Adam optimizer \cite{kingma2014adam} for 50 epochs.

\section{Evaluation and Results}\label{sec:results}

First, we evaluated all mapping function architectures to determine the top performing ones
using a random hold-out test set of 10K phrase--concept pairs.
Next, we tested the robustness and generalisability of the winning architectures on a new lexicon
consisting of 8.9K phrases from ICD-9-CM.
Finally, we tested the zero-shot learning capability of \ttc.
We did this by randomly selecting 1K concepts from our training set, removing all associated
3.4K phrase--concept pairs from the training dataset, and retraining the system.
The performance was evaluated by mapping the removed phrases to concepts that were never seen before.

In these tests we considered two performance metrics: accuracy and mean graph distance.
\begin{description}
  \item[Accuracy]
  is the proportion of test phrases that are mapped exactly to their corresponding concept (higher is better).
  This metric is often used for classification based approaches and is a standard metric for the mapping problem.

  \item[Mean graph distance]
  is the average graph distance (length of the shortest path) in the taxonomy between the target concept
  and the predicted concept (lower is better).
  This metric tells us \emph{how far} is the predicted concept from the exact match,
  and low scores are evidence for predictions that are immediate neighbours of the target concept,
  that is its children or parents.

\end{description}
Due to the probabilistic nature of node embeddings and the topology of our target graph,
it was possible that two distinct concepts were embedded to the exact same location in the node embedding space.
Since the mapping procedure involved nearest neighbour searches, we computed the performance metrics for the
top-$k$ results of these searches for $k=1,5,10,20,50$.
This also allowed us to better cover the test cases where multiple concepts were been assigned to a phrase.

\subsection{Intrinsic Evaluation}\label{sec:intrinsic}
We used a random hold-out test set of 10K phrase--concept pairs to find the top performing architectures.
The accuracy results are in Table~\ref{tab:snomed2snomed_accuracy}, and
the graph distance results are in Table~\ref{tab:snomed2snomed_graph_dist}.
The overall best system was the Bi-LSTM mapping function on top of FastText embeddings.
An interesting finding was that the cosine similarity yields consistently better accuracy scores,
while the $\ell_2$ distance provides for slightly smaller average graph distances.
\begin{table}[htbp]
  \centering\small
  \begin{tabular}{|c|c|c|c|c|c|c|c|}
    \hline
    \multicolumn{3}{|c|}{\ttc\ model} & \multicolumn{5}{c|}{Accuracy (top $k$)}                                                       \\
    Mapping & Word Emb & Metric   & $k=1$ & $k=5$ & $k=10$ & $k=20$ & $k=50$ \\
    \hline\hline
    \rowcolor{gray!25!white}
    Linear  & GloVe    & $\cos$   & 0.033 & 0.106 & 0.163  & 0.231  & 0.352  \\
    Linear  & GloVe    & $\ell_2$ & 0.014 & 0.062 & 0.099  & 0.153  & 0.245  \\
    \rowcolor{gray!25!white}
    Linear  & FastText & $\cos$   & 0.029 & 0.094 & 0.153  & 0.226  & 0.355  \\
    Linear  & FastText & $\ell_2$ & 0.012 & 0.050 & 0.091  & 0.143  & 0.237  \\
    \rowcolor{gray!25!white}
    CNN     & GloVe    & $\cos$   & 0.061 & 0.176 & 0.249  & 0.338  & 0.485  \\
    CNN     & GloVe    & $\ell_2$ & 0.045 & 0.150 & 0.214  & 0.294  & 0.419  \\
    \rowcolor{gray!25!white}
    CNN     & FastText & $\cos$   & 0.082 & 0.218 & 0.304  & 0.397  & 0.546  \\
    CNN     & FastText & $\ell_2$ & 0.067 & 0.181 & 0.256  & 0.350  & 0.495  \\
    \rowcolor{gray!25!white}
    Bi-LSTM & GloVe    & $\cos$   & 0.225 & 0.442 & 0.543  & 0.640  & 0.762  \\
    Bi-LSTM & GloVe    & $\ell_2$ & 0.195 & 0.402 & 0.497  & 0.602  & 0.722  \\
    \rowcolor{gray!25!white}
    Bi-LSTM & FastText & $\cos$   & \textbf{0.239} & \textbf{0.466} & \textbf{0.571}  & \textbf{0.671}  & \textbf{0.785}  \\
    Bi-LSTM & FastText & $\ell_2$ & 0.201 & 0.416 & 0.515  & 0.616  & 0.744  \\
    \hline
  \end{tabular}
  \caption{Intrinsic evaluation results: accuracy on a random hold-out test set of 10K phrase--concept pairs. }
  \label{tab:snomed2snomed_accuracy}
\end{table}
\begin{table}[htbp]
  \centering\small
  \begin{tabular}{|c|c|c|c|c|c|c|c|}
    \hline
    \multicolumn{3}{|c|}{\ttc\ model} & \multicolumn{5}{c|}{Mean graph distance (top $k$)}                                                       \\
    Mapping & Word Emb & Metric   & $k=1$ & $k=5$ & $k=10$ & $k=20$ & $k=50$ \\
    \hline\hline
    \rowcolor{gray!25!white}
    Linear  & GloVe    & $\cos$   & 2.08  & 2.23  & 2.30   & 2.35   & 2.42   \\
    Linear  & GloVe    & $\ell_2$ & 1.79  & 2.07  & 2.17   & 2.24   & 2.32   \\
    \rowcolor{gray!25!white}
    Linear  & FastText & $\cos$   & 2.01  & 2.17  & 2.23   & 2.28   & 2.35   \\
    Linear  & FastText & $\ell_2$ & 1.78  & 2.06  & 2.16   & 2.23   & 2.31   \\
    \rowcolor{gray!25!white}
    CNN     & GloVe    & $\cos$   & 1.71  & 1.95  & 2.03   & 2.11   & 2.19   \\
    CNN     & GloVe    & $\ell_2$ & 1.60  & 1.88  & 1.98   & 2.06   & 2.14   \\
    \rowcolor{gray!25!white}
    CNN     & FastText & $\cos$   & 1.60  & 1.86  & 1.95   & 2.03   & 2.13   \\
    CNN     & FastText & $\ell_2$ & 1.52  & 1.80  & 1.91   & 2.00   & 2.09   \\
    \rowcolor{gray!25!white}
    Bi-LSTM & GloVe    & $\cos$   & 1.28  & 1.63  & 1.76   & 1.88   & 2.01   \\
    Bi-LSTM & GloVe    & $\ell_2$ & 1.26  & 1.61  & 1.74   & 1.86   & \textbf{1.98}   \\
    \rowcolor{gray!25!white}
    Bi-LSTM & FastText & $\cos$   & 1.27  & 1.61  & 1.75   & 1.87   & 2.01   \\
    Bi-LSTM & FastText & $\ell_2$ & \textbf{1.25}  & \textbf{1.59}  & \textbf{1.73}   & \textbf{1.85}   & \textbf{1.98}   \\
    \hline
  \end{tabular}
  \caption{Intrinsic evaluation results: mean graph distance on a random hold-out test set of 10K phrase--concept pairs. }
  \label{tab:snomed2snomed_graph_dist}
\end{table}

The Bi-LSTM on FastText architecture achieved 23.9\% top-1 accuracy with the mean graph distance of 1.27.
This confirmed our expectation that the majority of cases that were not mapped to the exact target concept
were still mapped to one of its immediate neighbours.
The accuracy increased to 67.1\% when we considered the top-20 closest matches.

\subsection{Extrinsic Evaluation (ICD-9-CM)}

To check the robustness and generalisability of the \ttc\ system, we created an extrinsic evaluation task
consisting of 8.9K ICD-9-CM phrases mapped by medical experts from \cite{icd9snomed} to a unique \sno\ concept.
The lexicon of these phrases is different from our target taxonomy, because ICD-9-CM was developed for
assigning codes to diagnoses and procedures associated with hospital utilization in the United States.
The accuracy results are in Table~\ref{tab:icd2snomed_accuracy}, and
the graph distance results are in Table~\ref{tab:icd2snomed_graph_dist}.
\begin{table}[htbp]
  \centering\small
  \begin{tabular}{|c|c|c|c|c|c|c|c|}
    \hline
    \multicolumn{2}{|c|}{\ttc\ model} & \multicolumn{5}{c|}{Accuracy (top $k$)}                                                       \\
    Word Emb & Metric   & $k=1$ & $k=5$ & $k=10$ & $k=20$ & $k=50$ \\
    \hline\hline
    \rowcolor{gray!25!white}
    GloVe    & $\cos$   & 0.202 & 0.430 & 0.541  & 0.644  & 0.755  \\
    GloVe    & $\ell_2$ & 0.170 & 0.394 & 0.501  & 0.602  & 0.714  \\
    \rowcolor{gray!25!white}
    FastText & $\cos$   & \textbf{0.210} & \textbf{0.443} & \textbf{0.553}  & \textbf{0.658}  & \textbf{0.763}  \\
    FastText & $\ell_2$ & 0.181 & 0.403 & 0.507  & 0.614  & 0.725  \\
    \hline
  \end{tabular}
  \caption{Extrinsic evaluation: accuracy of the Bi-LSTM mapping function on 8.9K ICD-9-CM phrases. }
  \label{tab:icd2snomed_accuracy}
\end{table}
\begin{table}[htbp]
  \centering\small
  \begin{tabular}{|c|c|c|c|c|c|c|c|}
    \hline
    \multicolumn{2}{|c|}{\ttc\ model} & \multicolumn{5}{c|}{Mean graph distance (top $k$)}                                                       \\
    Word Emb & Metric   & $k=1$ & $k=5$ & $k=10$ & $k=20$ & $k=50$ \\
    \hline\hline
    \rowcolor{gray!25!white}
    GloVe    & $\cos$   & 1.42  & 1.74  & 1.88   & 2.00   & 2.14   \\
    GloVe    & $\ell_2$ & 1.40  & 1.72  & 1.85   & 1.97   & 2.10   \\
    \rowcolor{gray!25!white}
    FastText & $\cos$   & 1.40  & 1.73  & 1.87   & 1.99   & 2.14   \\
    FastText & $\ell_2$ & \textbf{1.36}  & \textbf{1.69}  & \textbf{1.84}   & \textbf{1.95}   & \textbf{2.09}   \\
    \hline
  \end{tabular}
  \caption{Extrinsic evaluation: mean graph distance of the Bi-LSTM mapping function on 8.9K ICD-9-CM phrases. }
  \label{tab:icd2snomed_graph_dist}
\end{table}

The Bi-LSTM model on FastText also yielded the best score in this experiment.
When using cosine similarity as similarity measurement,
it achieved 21\% in top-1 accuracy and 44\% in top-5 accuracy,
and when $\ell_2$ metric was used, it achieved 1.36 in graph distance.
The graph distance of 1.36 shows that, although the exact correct match was only 21\% at top-1 of \ttc\ ranking,
all the predicted concepts were close to the exact match in terms of taxonomy distance,
that is they are either synonyms, parents, or children of the exact match concept from 392K nodes.

This test set also allowed us to study the effect of extra knowledge of source phrases on mapping to the target taxonomy.
In our previous evaluations, we mapped phrases to one of the 392K concepts in the target taxonomy.
However, when we reduced the search space to 7.5K concepts that have at least one ICD-9-CM phrase mapped to them,
the accuracy and mean graph distance results improved significantly as shown in
Tables~\ref{tab:icd2snomed_restricted_accuracy} and~\ref{tab:icd2snomed_restricted_graph_dist}.
\begin{table}[htbp]
  \centering\small
  \begin{tabular}{|c|c|c|c|c|c|c|c|}
    \hline
    \multicolumn{2}{|c|}{\ttc\ model} & \multicolumn{5}{c|}{Accuracy (top $k$)}                                                       \\
    Word Emb & Metric   & $k=1$ & $k=5$ & $k=10$ & $k=20$ & $k=50$ \\
    \hline\hline
    \rowcolor{gray!25!white}
    GloVe    & $\cos$   & 0.379 & 0.698 & 0.793  & 0.855  & 0.905  \\
    GloVe    & $\ell_2$ & 0.350 & 0.655 & 0.754  & 0.822  & 0.876  \\
    \rowcolor{gray!25!white}
    FastText & $\cos$   & \textbf{0.389} & \textbf{0.710} & \textbf{0.804}  & \textbf{0.863}  & \textbf{0.914}  \\
    FastText & $\ell_2$ & 0.362 & 0.658 & 0.760  & 0.829  & 0.887  \\
    \hline
  \end{tabular}
  \caption{Extrinsic evaluation: accuracy of the Bi-LSTM mapping function on 8.9K ICD-9 phrases
  with the search space reduced to 7.5K concepts. }
  \label{tab:icd2snomed_restricted_accuracy}
\end{table}
\begin{table}[htbp]
  \centering\small
  \begin{tabular}{|c|c|c|c|c|c|c|c|}
    \hline
    \multicolumn{2}{|c|}{\ttc\ model} & \multicolumn{5}{c|}{Mean graph distance (top $k$)}                                                       \\
    Word Emb & Metric   & $k=1$ & $k=5$ & $k=10$ & $k=20$ & $k=50$ \\
    \hline\hline
    \rowcolor{gray!25!white}
    GloVe    & $\cos$   & 1.32  & 1.80  & 1.98   & 2.13   & 2.29   \\
    GloVe    & $\ell_2$ & 1.36  & 1.82  & 1.98   & 2.12   & 2.26   \\
    \rowcolor{gray!25!white}
    FastText & $\cos$   & \textbf{1.27}  & \textbf{1.78}  & 1.98   & 2.13   & 2.28   \\
    FastText & $\ell_2$ & 1.31  & 1.80  & \textbf{1.97}   & \textbf{2.11}   & \textbf{2.25}   \\
    \hline
  \end{tabular}
  \caption{Extrinsic evaluation: mean graph distance of the Bi-LSTM mapping function on 8.9K ICD-9 phrases
           with the search space reduced to 7.5K concepts. }
  \label{tab:icd2snomed_restricted_graph_dist}
\end{table}

\sno\ consists of a wide category of medical concepts including diseases, procedures, medications, and body structures
that are closely related, meaning that these nodes are close in the node embedding space.
Therefore, extra knowledge about the category of the taxonomy allowed us to reduce the search space
and boost the performance of \ttc.
In this test, the accuracy for top-1 almost doubled to 39\% and the graph distance went down to 1.27
by filtering the output concepts to be correlated to the ICD-9 concepts.

\subsection{Zero-shot Evaluation}
To evaluate the zero-shot learning capability,
we randomly selected 1K concepts from the taxonomy that appeared in our training set.
We then removed all 3.4K phrase--concept pairs associated to the selected concepts from the training set,
and used them as the test set.
We retrained the Bi-LSTM mapping network from scratch using the new training set, so that
all of the targets in the zero-shot test set were never seen before by the mapping function.
The accuracy and mean graph distance results are given in
Tables~\ref{tab:zeroshot_accuracy} and~\ref{tab:zeroshot_graph_dist}.
\begin{table}[htbp]
  \centering\small
  \begin{tabular}{|c|c|c|c|c|c|c|c|}
    \hline
    \multicolumn{2}{|c|}{\ttc\ model} & \multicolumn{5}{c|}{Accuracy (top $k$)}                                                       \\
    Word Emb & Metric   & $k=1$ & $k=5$ & $k=10$ & $k=20$ & $k=50$ \\
    \hline\hline
    \rowcolor{gray!25!white}
    GloVe    & $\cos$   & \textbf{0.242} & 0.449 & 0.538  & 0.622  & 0.737  \\
    GloVe    & $\ell_2$ & 0.227 & 0.391 & 0.481  & 0.583  & 0.706  \\
    \rowcolor{gray!25!white}
    FastText & $\cos$   & 0.221 & \textbf{0.499} & \textbf{0.591}  & \textbf{0.671}  & \textbf{0.780}  \\
    FastText & $\ell_2$ & 0.186 & 0.439 & 0.534  & 0.636  & 0.734  \\
    \hline
  \end{tabular}
  \caption{Zero-shot evaluation: accuracy of a newly trained Bi-LSTM mapping function on 3.4K phrases
           mapping to 1K previously unseen concepts. }
  \label{tab:zeroshot_accuracy}
\end{table}

\begin{table}[htbp]
  \centering\small
  \begin{tabular}{|c|c|c|c|c|c|c|c|}
    \hline
    \multicolumn{2}{|c|}{\ttc\ model} & \multicolumn{5}{c|}{Mean graph distance (top $k$)}                                                       \\
    Word Emb & Metric   & $k=1$ & $k=5$ & $k=10$ & $k=20$ & $k=50$ \\
    \hline\hline
    \rowcolor{gray!25!white}
    GloVe    & $\cos$   & 1.41  & 1.74  & 1.86   & 1.97   & 2.09   \\
    GloVe    & $\ell_2$ & \textbf{1.38}  & 1.70  & 1.82   & 1.93   & 2.04   \\
    \rowcolor{gray!25!white}
    FastText & $\cos$   & 1.41  & 1.69  & 1.81   & 1.92   & 2.05   \\
    FastText & $\ell_2$ & \textbf{1.38}  & \textbf{1.65}  & \textbf{1.78}   & \textbf{1.88}   & \textbf{2.00}   \\
    \hline
  \end{tabular}
  \caption{Zero-shot evaluation: mean graph distance of a newly trained Bi-LSTM mapping function on 3.4K phrases
  mapping to 1K previously unseen concepts. }
  \label{tab:zeroshot_graph_dist}
\end{table}

In this test, we were able achieve comparable results to the intrinsic task evaluations from Section~\ref{sec:intrinsic},
and we showed that our system is general enough to be able to map to unseen concepts in the training set.

%\subsection{Examples of cool mapped concepts}
%It is very interesting because ,.. In the case that the original phrase is a combination, we did something great or not that good.

\section{Discussion and Related Work}

\subsection{Discussion}
In this paper we have described our \ttc\ system for mapping arbitrary medical phrases to a large taxonomy.
In our system, we first used embedding models to project phrases in the source medical lexicon domain and
concept entities in in the target taxonomy domain into two different latent spaces.
We then learned a mapping function between those latent spaces
to associate points in the node embedding space to sequences of word embedding vectors.
We used nearest neighbour search in the target domain to find candidate medical concepts for a given phrase.
We tested a number of word embedding algorithms and mapping function architectures and
we obtained the best result using a Bi-LSTM mapping model and FastText word embeddings.

In summary, our experiments showed that \ttc\ is a robust and generalisable system that
achieved a satisfactory result at mapping ICD-9-CM Diagnosis phrases to \sno\,
and a comparable accuracy at zero-shot training.
More generally,
we confirmed our hypothesis that utilising the hierarchical structure of a biomedical taxonomy
to learn node representations can achieve a better mapping system between medical phrases and taxonomy concepts.

\subsection{Related work}

Concept mapping is a longstanding problem and there is a variety of existing approaches.
For example,
cTAKES \cite{savova2010mayo},
MetaMap \cite{aronson2010overview},
MedLEE \cite{friedman1994general},
KnowledgeMap \cite{denny2003knowledgemap}
NOBLE \cite{tseytlin2016noble}, and
PDD Graph \cite{wang2017pdd}
use annotated lookup tables in order to map clinical entities to concepts in controlled vocabularies
such as \sno and the Unified Medical Language System (UMLS).
To do so, they rely on symbolic natural language processing techniques, which makes them hard to generalise and scale.
cTAKES, was developed by Mayo Clinic and later transitioned to an Apache project.
MetaMap was developed by the National Library of Medicine for mapping biomedical text to the UMLS Metathesaurus.
MedLEE is one of the earliest clinical NLP systems developed and is mostly used for pharmacovigilance.

Word embeddings have been widely used in medical domain since they contain rich latent semantic information.
A few prominent examples include
\cite{moen2013distributional} that trained a set of word embeddings on PubMed data using Word2Vec;
\cite{liu2015effects} that studied drug name recognition utilizing word embeddings;
\cite{li2015using} that applied word embedding to extract events from the biomedical text; and
\cite{xu2015clinical} that examined the use of word embeddings for clinical abbreviation disambiguation.

% Medical entity/concept embedding
Recently researchers have started to explore the possibility of efficient representation learning in the medical domain.
\cite{choi2017gram} utilised a hierarchical structure from the medical term ontology by applying an attention mechanism.
\cite{choi2016medical, choi2016multi, choi2016learning} learned the representation of clinical codes using skip-gram
from structured visit records of patients.
\cite{tran2015learning} used a modified restricted RBM for representing EHR concepts
using a structured training process.
In a similar paper, \cite{lv2016clinical} used AEs to generate concept vectors from word-based concepts
extracted from clinical free text.

For cross-domain mapping in natural language processing,
\cite{artetxe2018generalizing} used linear transformation to improve bilingual word embedding mapping, and
\cite{artetxe2018robust} used adversarial training for cross-lingual word embeddings without parallel data.
Finally, in recommendation systems,
\cite{man2017cross} used a cross-domain recommendation system to leverage feedbacks or ratings from multiple domains
and improve recommendation performance.

\subsection{\ttc\ Applications}
\paragraph{NER}
The problem of named entity recognition (NER) is the task of extracting relevant concepts from free text.
Once extracted, such concepts need to be mapped to a taxonomy of known entities.
Our \ttc\ system can be used to solve this mapping problem.

\paragraph{Medical phrase representation}
Finding a representation for a phrase that can capture all the information in its sequence of words is
a challenging task in natural language processing.
The mapping function in our \ttc\ system can be seen as an encoding model
that is generating a representations for medical phrases.
Since these representations have been trained in a supervised manner using the information from the taxonomy,
they also represent their taxonomy position and the structure of their neighbourhood in the taxonomy.
Thus this supervised representation of the medical phrases can be used in different tasks
(i.e. medical outcome prediction) using transfer learning techniques.

\paragraph{Zero-shot learning}
A practical application of embedding mappings is the zero-shot transformation of concepts
that were missing in the training data.
Our system showed a satisfactory performance therefore it can be used for data pre-processing
when the training data is hard to obtain as is often the case.

\subsection{Future work}

Our system showed a satisfactory performance in mapping phrases between two taxonomies and
in a promising result towards solving the zero-shot problem.
More work can be done to make it better.

% \paragraph{Better word embedding or use sentence/paragraph embeddings}
% For word embedding step, there are recent work that does sentence/paragraph embeddings, such as ELMo, BERT, they can be used to improve word embedding by taking into account of the context of each phrase.

% \paragraph{Make it sensitivity to phrase grammar}
In EHRs, there are often phrases with a complex internal structure.
For example, a surgery code labelled \emph{Eye, Intra-ocular Foreign Body, Removal from Anterior Segment}
could be decomposed into three parts: location, problem, operation.
The performance of our system is yet to be evaluated by focusing on solely complex phrases.
On the other hand, some NLP features such as part of speech can be coupled into our system to solve this problem.

% \paragraph{Improve taxonomy incompleteness}
In the taxonomy, there exist nodes that have the exact same connections to the rest of the taxonomy.
Therefore, they have the same representation in our node embedding space.
Our current system does not address this issue and these nodes will not differentiated.
This problem can be solved by adding extra informative connections to those nodes to make their neighbourhoods different.
It is also possible to use two taxonomies to train the node embeddings,
to differentiate these nodes and yield better performance.

\bibliographystyle{hplain}
{\footnotesize \bibliography{t2c}}

\end{document}